\documentclass[letterpaper, 10 pt, conference]{ieeeconf}  
\IEEEoverridecommandlockouts                             
\usepackage{cite}
\usepackage{amsmath,amssymb,amsfonts}
\usepackage{algorithmic}
\usepackage{graphicx}
\usepackage{textcomp}
\usepackage{tabularx}
\usepackage{siunitx}
\usepackage{xcolor}
\usepackage{balance}
\usepackage{bm}
\usepackage[ruled,linesnumbered]{algorithm2e}
\usepackage{subcaption} 
\usepackage{multirow}
\usepackage{threeparttable}

\title{\LARGE \bf
GeoPro-VO: Dynamic Obstacle Avoidance with Geometric Projector Based on Velocity Obstacle 
}

\author{Jihao Huang$^{1}$, Xuemin Chi$^{1}$, Jun Zeng$^{2}$, Zhitao Liu$^{1\dagger}$, Hongye Su$^{1}$%
\thanks{This work was supported in part by National Key R\&D Program of China (Grant NO. 2021YFB3301000); National Natural Science Foundation of China (NSFC:62173297), Zhejiang Key R\&D Program (Grant NO. 2022C01035).}%
\thanks{$^{1}$Jihao Huang, Xuemin Chi, Zhitao Liu and Hongye Su are with the State Key Laboratory of Industrial Control Technology, Institute of Cyber-Systems and Control, Zhejiang University, Hangzhou, China {\tt\footnotesize \{jihaoh, chixuemin, ztliu, hysu\}@zju.edu.cn}.
$^{2} $Jun Zeng is with Cruise LLC, USA {\tt\footnotesize jun.zeng@getcruise.com}.
}%
\thanks{$^\dagger$ Corresponding author.} %
}

\begin{document}
\maketitle
\thispagestyle{empty}
\pagestyle{empty}

\begin{abstract}
Optimization-based approaches are widely employed to generate optimal robot motions while considering various constraints, such as robot dynamics, collision avoidance, and physical limitations.
It is crucial to efficiently solve the optimization problems in practice, yet achieving rapid computations remains a great challenge for optimization-based approaches with nonlinear constraints.
In this paper, we propose a geometric projector for dynamic obstacle avoidance based on velocity obstacle (GeoPro-VO) by leveraging the projection feature of the velocity cone set represented by VO.
Furthermore, with the proposed GeoPro-VO and the augmented Lagrangian spectral projected gradient descent (ALSPG) algorithm, we transform an initial mixed integer nonlinear programming problem (MINLP) in the form of constrained model predictive control (MPC) into a sub-optimization problem and solve it efficiently.
Numerical simulations are conducted to validate the fast computing speed of our approach and its capability for reliable dynamic obstacle avoidance.
\end{abstract}

\section{Introduction}
\subsection{Motivation}
Most robot motion planning tasks can be formulated as optimization problems, with various constraints like robot dynamics, safety and physical limitations~\cite{wang2022geometrically}.
Nonlinear model predictive control (NMPC) can consider all these constraints within a short prediction horizon and solve the constrained optimization problem in real time~\cite{wang2024hierarchical}.
However, when the prediction horizon is long and non convex constraints are present, real time solution can not be guaranteed.
In this paper, we propose to construct the geometric projector for dynamic obstacle avoidance based on velocity obstacle (GeoPro-VO) and integrate it with the augmented Lagrangian spectral projected gradient descent (ALSPG) algorithm to reformulate the nominal constrained NMPC problem into a sub-optimization problem for efficiently solving.

\subsection{Related Works}
\subsubsection{Projection-based Optimization Approaches}
In most optimization problems, constraints such as safety and physical limitations are typically described using geometric set primitives or combinations thereof. 
For instance, physical limitations are commonly modeled as box constraints.
Safety constraints, like avoiding collisions with ellipsoids or polytopes, are often expressed through quadratic sets, hyperplanes, and cone constraints.
Some studies propose to leverage the projection feature of these sets instead of directly integrating them as geometric constraints in optimization problems.
The projected gradient descent algorithm~\cite{torrisi2018projected} is utilized to solve the sub-optimization problem of sequential quadratic programming (SQP) with faster efficiency than the second-order solver SNOPT~\cite{gill2005snopt}.
Spectral projected gradient descent (SPG)~\cite{birgin2009spectral} improves on the previous approach by utilizing curvature information through spectral step sizes.
It has been applied across various fields for its great practical performance~\cite{birgin2014spectral}, outperforming even commercially available second-order solvers if a computationally efficient projection is used.
However, using SPG alone can only solve optimization problems with physical constraints on the control's boundaries, it is not sufficient to solve optimization problems in robotics with complex nonlinear constraints, such as collision avoidance constraints.

In addition, some studies~\cite{andreani2008augmented, birgin2014practical, jia2023augmented, girgin2023projection} integrate SPG with the augmented Lagrangian framework to overcome the limitations of SPG, and this algorithm is commonly referred to as augmented Lagrangian spectral projected gradient descent (ALSPG).
ALSPG first incorporates the constraint violation size as a cost function into the objective function of the nominal optimization problem, resulting in a sub-optimization problem only with physical constraints on control boundaries.
In addition, the sub-optimization problem is solved by SPG efficiently.
To effectively reformulate the nominal optimization problem with the augmented Lagrangian framework, a more general framework based on the concept of geometric projector (GeoPro) is proposed in~\cite{chi2023geometric}.
GeoPro can effectively project a point onto a set, enabling us to quickly calculate the magnitude of constraint violations.
Chi~\emph{et al.}~\cite{chi2023geometric} briefly describes collision avoidance constraints using GeoPro as follows:
if the current state is within the set of obstacle, GeoPro will efficiently project the state to the set's boundary;
otherwise, it will maintain the current state.
However, \cite{chi2023geometric} mainly constructs GeoPro for collision avoidance based on the Euclidean distance and projects the unsafe robot positions to the boundary of obstacles, which may not perform well with dynamic obstacles.

\subsubsection{Velocity Obstacle}
Collision cone~\cite{Chakravarthy1998} is a commonly crucial concept for local and reactive collision avoidance, and velocity obstacle (VO) is indeed a specific type of collision cone.
VO and its variants~\cite{fiorini1998motion, van2008reciprocal, berg2011reciprocal, snape2011hybrid} have been widely employed to realize collision avoidance and navigation with both static and dynamic obstacles for robots.
VO is defined as the set of all velocities for a robot which will lead to a collision with an obstacle at some future moment, with the assumption that the obstacle maintains its current velocity.
By choosing a velocity outside of VO induced by the obstacle, collision avoidance is guaranteed between them.
In addition, the robot can select a velocity which is outside of any VO induced by all obstacles to avoid collisions with all obstacles.
Since VO explicitly considers the velocity of the obstacle and described the sets of unsafe velocity, making it suitable for dynamic obstacle avoidance.
In addition, in order to design dynamic collision avoidance controllers for the robot controlled by acceleration, some studies~\cite{Cheng2017decentralized, zhang2022velocity} formulate collision avoidance constraints based on VO or its variants.
Zhang~\emph{et al.}~\cite{zhang2022velocity} propose integrating VO with the nonlinear model predictive control (NMPC), and formulate the obstacle avoidance constraint for velocity outside of VO as the velocity should be within the union of two half spaces.
Since the condition that the velocity is within the union of two half spaces cannot be represent directly in terms of one single constraint, it is represented as a form in which at least one of two individual constraints is required to be met (each constraint requires the velocity within a half space).
With such constraints, the nominal constrained NMPC problem introduces integer variables and becomes a mixed integer nonlinear programming problem (MINLP)~\cite{sutradhar2016minlp}, which requires a lot of time to solve.
However, the velocity set represented by VO is a cone with good projection feature that can be integrated with GeoPro.
Therefore, we aim to construct GeoPro for dynamic obstacle avoidance based on VO as follows:
if the robot's velocity is within VO, project it to the boundary of VO; 
otherwise, maintain it.
With the introduction of GeoPro, the solution of the initial MINLP can be avoided and the overall solution efficiency can be improved.

\subsection{Contributions}
In this paper, we propose to construct GeoPro for dynamic obstacle avoidance based on VO (GeoPro-VO) and solve the initial optimization problem with GeoPro-VO and ALSPG efficiently.
The key contributions are as follows:
\begin{itemize}
    \item We leverage the projection feature of the velocity cone set represented by VO to construct GeoPro-VO, which projects the unsafe velocity inside the VO outside the VO to achieve dynamic obstacle avoidance.
    \item By combining GeoPro-VO and the ALSPG algorithm, the nominal MINLP in the form of constrained NMPC is reformulated as a sub-optimization problem only with the physical constraints on controls, which can be solved by SPG with computational efficiency.
    \item Extensive numerical simulations have been conducted to validate that our approach can efficiently solve the sub-optimization problem and improve navigation safety. With proposed approach, the robot is able to reach its destination while avoiding collisions with both static and dynamic obstacles, even when the prediction horizon of MPC is short.
\end{itemize}

\subsection{Organization}
The rest of this paper are organized as follows:
we formally define the nominal constrained NMPC optimization problem, review the concept of GeoPro and VO in Sec.~\ref{sec:problem_define}.
In Sec.~\ref{sec:method}, we introduce how to construct GeoPro-VO and integrate it with the ALSPG algorithm.
The effectiveness and performance of our approach are demonstrated in Sec.~\ref{sec:experiments}.
Sec.~\ref{sec:con} concludes the paper.
\section{Problem Definition \& Preliminaries}
\label{sec:problem_define}
\subsection{Problem Definition}
Assume there is a robot with multiple obstacles in the environment.
The robot's dynamics follow the double integral model:
\begin{equation}
    \dot{x} = v_\text{x}, \dot{y} = v_\text{y}, \dot{v_\text{x}} = a_\text{x}, \dot{v_\text{y}} = a_\text{y},
    \label{eq:robot_dynamics}
\end{equation}
where the robot states are $\bm{x} = [x, y, v_\text{x}, v_\text{y}]^T$ with controls $\bm{u} = [a_\text{x}, a_\text{y}]^T$.
The obstacle dynamics are similar to those of the robot with states $\bm{x}_{\text{o}_i} = [x_\text{o}, y_\text{o}, v_{\text{ox}}, v_\text{oy}]^T$. 
The positions and velocities of the robot and obstacle are represented by $\bm{p}_\text{R} = [x, y]^T$, $\bm{p}_{\text{o}_i} = [x_\text{o}, y_\text{o}]^T$, $\bm{v}_\text{R} = [v_\text{x}, v_\text{y}]^T$ and $\bm{v}_{\text{o}_i} = [v_\text{ox}, v_\text{oy}]^T$.
Let $\bm{p}_\text{R}^{\text{o}_i} := \bm{p}_\text{R} - \bm{p}_{\text{o}_i}$ and $\bm{v}_\text{R}^{\text{o}_i} := \bm{v}_\text{R} - \bm{v}_{\text{o}_i}$ be the relative position and velocity between $\text{R}$ and $\text{o}_i$.
Both robot and obstacles are assumed to be circular-shaped with radii $r$ and $r_\text{o}$, respectively.

Our work focuses on navigating a robot to its destination while avoiding collisions with all obstacles.
The entire problem is formulated as a constrained nonlinear model predictive control (NMPC) optimization problem:

\noindent\rule{\columnwidth}{0.8pt}
\textbf{NMPC:}
\begin{subequations}
\begin{align}
    \min_{\mathcal{X}, \, \mathcal{U}} \, \, & l(\mathcal{X}, \mathcal{U}) = \sum_{k=0}^{N-1} h(\bm{x}_k, \bm{u}_k) \\
    \text{s.t.} ~& \bm{x}_{k+1} = f(\bm{x}_k, \bm{u}_k), \label{eq:system_cons} \\
    & \bm{g}_i(\bm{x}_k) \in \mathcal{C}_i, \forall i = 1,2, \dots, N_p, \label{eq:collision_cons} \\
    & \bm{x} \in \mathcal{D}_x, \label{eq:states_cons} \\
    & \bm{u} \in \mathcal{D}_u, \label{eq:controls_cons}
\end{align}
\label{eq:nmpc}
\end{subequations}
\noindent\rule{\columnwidth}{0.8pt}
\noindent
where $\mathcal{X} = [\bm{x}_1^T, \dots, \bm{x}_N^T]^T \in \mathbb{R}^{N \cdot n_x}$, $N$ is the predict horizon of NMPC, $\mathcal{U} = [\bm{u}_0^T, \dots, \bm{u}_{N-1}^T]^T \in \mathbb{R}^{N \cdot n_u}$ and $N_p$ is the number of safety constraints.
$h: \mathbb{R}^{n_x} \times \mathbb{R}^{n_u} \to \mathbb{R}$ is the cost function which depends on the robot's task, such as navigating robot to the goal position or minimizing controls.
The system dynamics $f: \mathbb{R}^{n_x} \times \mathbb{R}^{n_u} \to \mathbb{R}^{n_x}$ are same to \eqref{eq:robot_dynamics}.
$\bm{g}_i(\bm{x}_k) \in \mathcal{C}_i$ means the robot should within the security zone, where $\bm{g}_i: \mathbb{R}^{n_x} \to \mathbb{R}^{n_i}, \mathcal{C}_i \in \mathbb{R}^{n_i}$.
Constraints \eqref{eq:states_cons} and \eqref{eq:controls_cons} represent the physical limitations on states and controls respectively, indicating that both states and controls should fall within a reasonable range.

\subsection{Geometric Projector}
Geometric projector (GeoPro)~\cite{chi2023geometric} is proposed to project geometric constraints and then reformulated the nominal optimization problem to a sub-optimization problem with the augmented Lagrangian method.
Taking the nominal constrained NMPC problem \eqref{eq:nmpc} as an example (\eqref{eq:states_cons} is omitted here for brevity), it can be reformulated with the classic augmented Lagrangian approach as
\begin{equation*}
    \mathcal{L}(\mathcal{X}, \mathcal{U}, \bm{\lambda}, \bm{\rho}) := l(\mathcal{X}, \mathcal{U}) + \sum_{i=1}^{N_p} \frac{\rho_{\mathcal{C}_i}}{2} d_{\mathcal{C}_i}^2(\bm{g}_i(\mathcal{X}) + \frac{\bm{\lambda}_{\mathcal{C}_i}}{\rho_{\mathcal{C}_i}}),
    \label{eq:lagrangian1}
\end{equation*}
where $\bm{\lambda}=[\bm{\lambda}_{\mathcal{C}_1}, \dots, \bm{\lambda}_{\mathcal{C}_{N_p}}]^T$, $\bm{\lambda}_{\mathcal{C}_i} \in \mathbb{R}^{n_i}$ is the Lagrangian multipliers, $\bm{\rho} = [\rho_{\mathcal{C}_1}, \dots, \rho_{\mathcal{C}_{N_p}}]^T$, $\rho_{\mathcal{C}_i} \in \mathbb{R}$ is the penalty parameters and $d_{\mathcal{C}_i}^2(\bm{y}_0) = \min_{\bm{y}} \| \bm{y} - \bm{y}_0 \|, \bm{y} \in \mathcal{C}_i$, which means the minimum distance from $\bm{y}_0$ to the set $\mathcal{C}_i$.
GeoPro is defined as $\mathcal{P}_{\mathcal{C}_i}(\bm{y}_0) = \arg\min_{\bm{y} \in \mathcal{C}_i} \| \bm{y} - \bm{y}_0 \|$.

In addition, we can use the GeoPro $\mathcal{P}_{\mathcal{C}_i}$ to obtain the following compact form:
\begin{equation}
\begin{aligned}
    \mathcal{L}(& \mathcal{X}, \mathcal{U}, \bm{\lambda}, \bm{\rho}) := l(\mathcal{X}, \mathcal{U}) + \\
    & \sum_{i=1}^{N_p} \frac{\rho_{\mathcal{C}_i}}{2} \| \bm{g}_i(\mathcal{X}) + \frac{\bm{\lambda}_{\mathcal{C}_i}}{\rho_{\mathcal{C}_i}} - \mathcal{P}_{\mathcal{C}_i}(\bm{g}_i(\bm{x}) + \frac{\bm{\lambda}_{\mathcal{C}_i}}{\rho_{\mathcal{C}_i}}) \|^2,
\end{aligned}
\label{eq:lagrangian2}
\end{equation}
where the level of constraints illegality are incorporated into the objective function as additional costs through GeoPro.
If $\bm{g}_i(\bm{x}_k) \in \mathcal{C}_i$, no additional cost is incurred in \eqref{eq:lagrangian2} since $\mathcal{P}_{\mathcal{C}_i}(\bm{g}_i(\bm{x}_k)) = \bm{g}_i(\bm{x}_k)$ retains the original state.
Conversely, if $\bm{g}_i(\bm{x}_k) \notin \mathcal{C}_i$, we need to project $\bm{g}_i(\bm{x}_k)$ with GeoPro onto the surface of $\mathcal{C}_i$ with additional cost in \eqref{eq:lagrangian2}.
Using obstacle avoidance constraints as examples, as both the robot and obstacles are in circular-shaped, we can establish obstacle avoidance constraints using the Euclidean distance as follows:
\begin{equation}
    \| \bm{p}_\text{R}^{\text{o}_i} \|_2 \geq r + r_{\text{o}_i}.
    \label{eq:Euclidean}
\end{equation}
In addition, we can construct the GeoPro based on the constraint~\eqref{eq:Euclidean} mentioned above, referred to as GeoPro-ED:
\begin{equation}
\resizebox{\columnwidth}{!}{$
    \mathcal{P}_{\text{ED}_i}(\bm{p}_\text{R}) = 
    \left\{
    \begin{aligned}
        & \bm{p}_\text{R}, \| \bm{p}_\text{R}^{\text{o}_i} \|_2 \geq r + r_{\text{o}_i} \\
        & \bm{p}_\text{R} + \frac{\bm{p}_\text{R}^{\text{o}_i} (r + r_{\text{o}_i} - \| \bm{p}_\text{R}^{\text{o}_i} \|_2)}{\| \bm{p}_\text{R}^{\text{o}_i} \|_2},  \| \bm{p}_\text{R}^{\text{o}_i} \|_2 < r + r_{\text{o}_i}
    \end{aligned}
    \right..
$}
\label{eq:Euclid_pro}
\end{equation}
Furthermore, GeoPro can also consider box constraints like \eqref{eq:states_cons} to restrict the robot's states.
To conclusion, GeoPro plays a significant role when constraints are violated.

\subsection{Velocity Obstacle}
\label{subsec:vo}
\begin{figure} 
    \centering
    \includegraphics[width=0.95\linewidth]{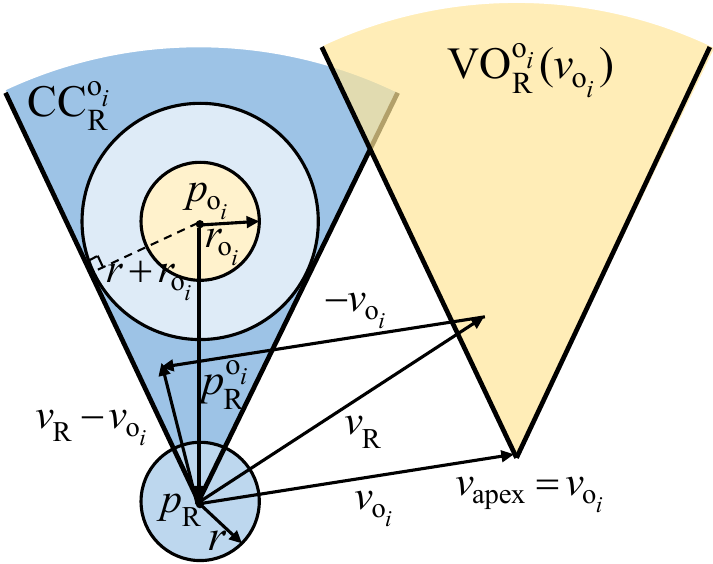}
    \caption{Velocity obstacle $\text{VO}_\text{R}^{\text{o}_i}(\bm{v}_{\text{o}_i})$ of robot $\text{R}$ induced by the obstacle $\text{o}_i$. $\text{CC}_\text{R}^{\text{o}_i}$ denotes the collision cone between them. If the relative velocity $\bm{v}_\text{R} -\bm{v}_{\text{o}_i} \in \text{CC}_\text{R}^{\text{o}_i}$ or the robot velocity $\bm{v}_\text{R} \in \text{VO}_\text{R}^{\text{o}_i}(\bm{v}_{\text{o}_i})$, a collision will occur between $\text{R}$ and $\text{o}_i$.}
    \label{fig:vo_explain}
\end{figure}
The subscript $k$ of time notation is dropped for simplicity.
Let there be a robot $\text{R}$ and an obstacle $\text{o}_i$ with radii $r$ and $r_{\text{o}_i}$.
The velocity obstacle (VO) for $\text{R}$ induced by $\text{o}_i$ is denoted as $\text{VO}_\text{R}^{\text{o}_i}(\bm{v}_{\text{o}_i})$, which includes all velocities of $\text{R}$ that would lead to a collision between $\text{R}$ and $\text{o}_i$ at future time moment, with assumption that $\text{o}_i$ moves with the constant velocity $\bm{v}_{\text{o}_i}$.

Before introducing VO, let $A \oplus B = \{a + b \big| a \in A, b \in B\}$ be the Minkowski sum of sets $A$ and $B$, and $\lambda(\bm{p},\bm{v}) = \{\bm{p} + t\bm{v} \big| t>0 \}$ denotes a ray starting at point $\bm{p}$ and in the direction of vector $\bm{v}$.
If a ray starting at $\bm{p}_\text{R}$ and heading in the direction of the relative velocity $\bm{v}_\text{R} - \bm{v}_{\text{o}_i}$ intersects the Minkowski sum of $\text{o}_i$ and $-\text{R}$ centered at $\bm{p}_{\text{o}_i}$, then $\bm{v}_\text{R} \in \text{VO}_\text{R}^{\text{o}_i}(\bm{v}_{\text{o}_i})$.
Hence, $\text{VO}_\text{R}^{\text{o}_i}(\bm{v}_{\text{o}_i})$ is defined as follows~\cite{fiorini1998motion}:
\begin{equation*}
    \text{VO}_\text{R}^{\text{o}_i}(\bm{v}_{\text{o}_i}) = \big\{
    \bm{v}_\text{R} \big| \lambda(\bm{p}_\text{R}, \bm{v}_\text{R} - \bm{v}_{\text{o}_i}) \cap \text{o}_i \oplus -\text{R} \neq \emptyset 
    \big\}.
\label{eq:vo_define}
\end{equation*}
Therefore, if $\bm{v}_\text{R} \in \text{VO}_\text{R}^{\text{o}_i}(\bm{v}_{\text{o}_i})$, $\text{R}$ and $\text{o}_i$ will collide at some future moment.
On the contrary, if $\bm{v}_\text{R} \notin \text{VO}_\text{R}^{\text{o}_i}(\bm{v}_{\text{o}_i})$, $\text{R}$ and $\text{o}_i$ will never collide.
In fact, $\text{VO}_\text{R}^{\text{o}_i}(\bm{v}_{\text{o}_i})$ is a cone with its apex at $\bm{v}_{\text{o}_i}$, as shown in Fig.~\ref{fig:vo_explain}, and it can be translated from the collision cone (CC) defined as follows:
\begin{equation*}
    \text{CC}_\text{R}^{\text{o}_i} = \big\{\bm{v}_\text{R}-\bm{v}_{\text{o}_i} \big| \lambda(\bm{p}_\text{R}, \bm{v}_\text{R} -\bm{v}_{\text{o}_i}) \cap \text{o}_i \oplus -\text{R} \neq \emptyset 
    \big\}.
\label{eq:cc_define}
\end{equation*}
In addition, $\text{VO}_\text{R}^{\text{o}_i}(\bm{v}_{\text{o}_i}) = \text{CC}_\text{R}^{\text{o}_i} \oplus \bm{v}_{\text{o}_i}$, the only distinction between these two cones lies in their apex positions: $\text{CC}_\text{R}^{\text{o}_i}$ is at $\bm{0}$ as it considers relative velocity.

VO is commonly utilized for enabling robot to avoid collisions with obstacles by choosing a velocity that lies outside any of the VO induced by each obstacle.
Many works have been proposed to address the limitations of VO in effectively avoiding collisions between multiple robots, such as reciprocal velocity obstacle (RVO)~\cite{van2008reciprocal}, hybrid reciprocal velocity obstacle (HRVO)~\cite{snape2011hybrid} and optimal reciprocal collision avoidance (ORCA)~\cite{berg2011reciprocal}.
Since VO defines the cone set of velocity, we aim to take full advantage of the projection properties of the cone set to construct GeoPro, more details can refer to Sec.~\ref{subsec:GeoPro_vo}.

\section{Method Design}
\label{sec:method}
In this section, we will first demonstrate how to build the GeoPro for dynamic obstacle avoidance based on VO (GeoPro-VO) in the velocity space.
Additionally, we will discuss integrating the GeoPro-VO into ALSPG and provide a more detailed overview of the algorithmic steps involved in ALSPG.

\subsection{Geometric Projector Based on Velocity Obstacle}
\label{subsec:GeoPro_vo}
\begin{figure} 
    \centering
    \includegraphics[width=0.95\linewidth]{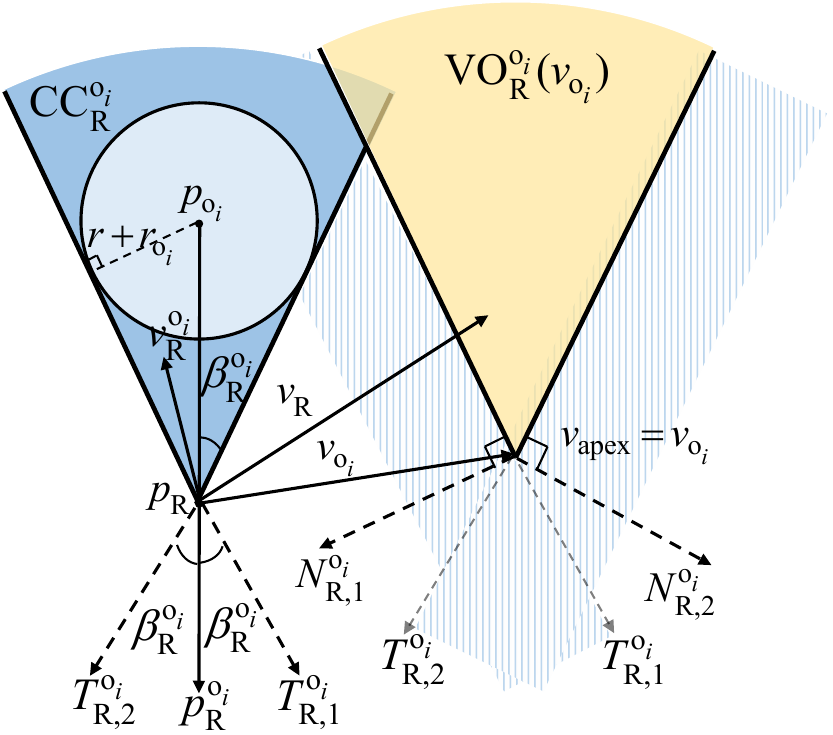}
    \caption{Constructing the hyperplanes for $\text{VO}_\text{R}^{\text{o}_i}(\bm{v}_{\text{o}_i})$. $\text{R}$ can avoid collisions with $\text{o}_i$ if $\bigcup_{m \in \{1, 2\}} (\bm{N}_{\text{R},m}^{\text{o}_i})^T \bm{v}_\text{R} \geq c_{\text{R}, m}^{\text{o}_i}$, which means $\bm{v}_\text{R} \notin \text{VO}_\text{R}^{\text{o}_i}(\bm{v}_{\text{o}_i})$.}
    \label{fig:vo_hyper}
\end{figure}
In this section, we mainly introduce how to construct the GeoPro-VO based on the robot dynamics~\eqref{eq:robot_dynamics}.
For the robot dynamics without horizontal and vertical velocity $v_\text{x}$ and $v_\text{y}$, we can use the differentiation of horizontal position and vertical position with respect to time to represent $v_\text{x}$ and $v_\text{y}$.
Taking the bicycle model as an example:
\begin{equation*}
    v_\text{x} = \dot{x} = v\cos\theta, v_\text{y} = \dot{y} = v\sin\theta.
\end{equation*}
In the following, we will demonstrate how to leverage the project feature of the cone set to construct GeoPro-VO.

As we mentioned above, the robot $\text{R}$ will avoid collisions with the obstacle $\text{o}_i$ when $\bm{v}_\text{R} \notin \text{VO}_\text{R}^{\text{o}_i}(\bm{v}_{\text{o}_i})$.
The two tangent vectors $\bm{T}_{\text{R}, 1}^{\text{o}_i}$ and $\bm{T}_{\text{R}, 2}^{\text{o}_i}$ of $\text{CC}_\text{R}^{\text{o}_i}$ shown in Fig.~\ref{fig:vo_hyper} can be computed as 
\begin{equation*}
    \bm{T}_{\text{R}, 1}^{\text{o}_i} = \bm{R}( \beta_{\text{R}}^{\text{o}_i}) \bm{p}_{\text{R}}^{\text{o}_i}, \,\,
    \bm{T}_{\text{R}, 2}^{\text{o}_i} = \bm{R}(-\beta_{\text{R}}^{\text{o}_i}) \bm{p}_{\text{R}}^{\text{o}_i},
    \label{eq:tangent_vectors}
\end{equation*}
where 
\begin{equation*}
    \bm{R}(\beta) = 
    \left[\begin{array}{cc}
    \cos\beta & -\sin\beta \\
    \sin\beta &  \cos\beta
    \end{array}\right].
\end{equation*}
Furthermore, the normal vectors of these two tangent vectors can be obtained by rotating them like
\begin{equation*}
    \bm{N}_{\text{R}, 1}^{\text{o}_i} = \bm{R}(-\frac{\pi}{2}) \bm{T}_{\text{R}, 1}^{\text{o}_i}, \, \,
    \bm{N}_{\text{R}, 2}^{\text{o}_i} = \bm{R}( \frac{\pi}{2}) \bm{T}_{\text{R}, 2}^{\text{o}_i}.
    \label{eq:normal_vectors}
\end{equation*}
Since VO is translated form CC, the normal and tangent vectors of VO are the same as these of CC, as shown in Fig.~\ref{fig:vo_hyper}.
With the property $\text{VO}_\text{R}^{\text{o}_i}(\bm{0}) = \text{CC}_\text{R}^{\text{o}_i}$, we can derive if $\bm{v}_\text{R}^{\text{o}_i} \notin \text{CC}_\text{R}^{\text{o}_i}$, collision avoidance is achieved.
This condition can be represented by the following two linear constraints:
\begin{equation}
    \bm{v}_\text{R}^{\text{o}_i} \notin \text{CC}_\text{R}^{\text{o}_i} \iff \bigcup_{m \in \{1, 2\}} (\bm{N}_{\text{R},m}^{\text{o}_i})^T \bm{v}_\text{R}^{\text{o}_i} \geq 0,
    \label{eq:vo_cons1}
\end{equation}
which means we can represent the safe region of $\bm{v}_\text{R}^{\text{o}_i}$ by using the union of two half spaces.
In addition, we reformulate condition \eqref{eq:vo_cons1} with the absolute velocity as 
\begin{equation}
    \bm{v}_\text{R} \notin \text{VO}_\text{R}^{\text{o}_i}(\bm{v}_{\text{o}_i}) \iff  \bigcup_{m \in \{1, 2\}} (\bm{N}_{\text{R},m}^{\text{o}_i})^T \bm{v}_\text{R} \geq c_{\text{R}, m}^{\text{o}_i},
    \label{eq:vo_cons2}
\end{equation}
where $c_{\text{R}, m}^{\text{o}_i} = (\bm{N}_{\text{R}, m}^{\text{o}_i})^T \bm{v}_{\text{o}_i}$ is the scalar corresponding to the $m^{\text{th}}$ linear constraint of $\bm{v}_\text{R} \notin \text{VO}_\text{R}^{\text{o}_i}(\bm{v}_{\text{o}_i})$.
Condition \eqref{eq:vo_cons2} is precisely the obstacle avoidance constraint based on VO.
However, when we attempt to directly incorporate \eqref{eq:vo_cons2} into \eqref{eq:nmpc}, integer variables are required to ensure that at least one of these two constraints is required to be met.
The detailed form of VO-based NMPC (VO-NMPC) is as follows:

\noindent\rule{\columnwidth}{0.8pt}
\textbf{VO-NMPC:}
\begin{subequations}
\begin{align}
    \min_{\mathcal{X}, \, \mathcal{U}} \, \, & l(\mathcal{X}, \mathcal{U}) = \sum_{k=0}^{N-1} h(\bm{x}_k, \bm{u}_k) \notag  \\
    \text{s.t.} ~& \eqref{eq:system_cons}, \eqref{eq:states_cons}, \eqref{eq:controls_cons}, \notag \\
    & c_{\text{R}, m}^{\text{o}_i} - (\bm{N}_{\text{R},m}^{\text{o}_i})^T \bm{v}_\text{R} \leq G(1 - z_{i, m}), \notag \label{eq:integer_cons1} \\
    & \qquad \quad \; \; i = 1,2, \dots, N_p, m = 1, 2, \\
    & \sum_{m=1}^{2} z_{i, m} \geq 1, z_{i, m} \in \{0, 1\}. \label{eq:integer_cons2} 
\end{align}
\label{eq:vo_mpc}
\end{subequations}
\noindent\rule{\columnwidth}{0.8pt}
\noindent
where $z_{i, m}$ is the integer variable and $G$ is a large positive real number, such as $10^5$.
The subscripts $k$ for time notation of \eqref{eq:integer_cons1} and \eqref{eq:integer_cons2} are also dropped for brevity.
The nominal optimization problem~\eqref{eq:nmpc} is transformed into a mixed integer nonlinear programming problem (MINLP).
Solving MINLP requires a large amount of computation time, making it unsuitable for real time applications.

\begin{figure} 
    \centering
    \includegraphics[width=0.75\linewidth]{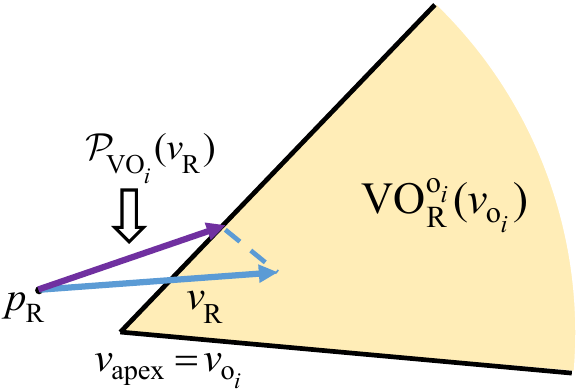}
    \caption{If $\bm{v}_\text{R} \in \text{VO}_\text{R}^{\text{o}_i}(\bm{v}_{\text{o}_i})$, project the unsafe velocity to the nearest hyperplane.}
    \label{fig:vo_project}
\end{figure}

In our work, we aim to address the aforementioned issues using GeoPro-VO and ALSPG.
If $\bm{v}_\text{R} \notin \text{VO}_\text{R}^{\text{o}_i}(\bm{v}_{\text{o}_i})$, GeoPro-VO will maintain the original state as the velocity is within the safe range, i.e., $\mathcal{P}_{\text{VO}_i}(\bm{v}_\text{R}) = \bm{v}_\text{R}$, where $\mathcal{P}_{\text {VO}_i}$ also falls under $\mathcal {P}_{\mathcal {C}_i}$.
Moreover, if $\bm{v}_\text{R} \in \text{VO}_\text{R}^{\text{o}_i}(\bm{v}_{\text{o}_i})$, which can be represented using the two linear constraints as:
\begin{equation}
    \bm{v}_\text{R} \in \text{VO}_\text{R}^{\text{o}_i}(\bm{v}_{\text{o}_i}) \iff \bigcap_{m \in \{1, 2\}} (\bm{N}_{\text{R}, m}^{\text{o}_i})^T \bm{v}_\text{R} \leq c_{\text{R}, m}^{\text{o}_i}.
    \label{eq:vo_cons3}
\end{equation}
Rewriting condition \eqref{eq:vo_cons3} to a matrix form as 
\begin{equation*}
\begin{gathered}
    \bm{v}_\text{R} \in \text{VO}_\text{R}^{\text{o}_i}(\bm{v}_{\text{o}_i}) \iff \bm{v}_\text{R} \in \mathcal{C}_i^\text{VO} := \{ \bm{v} \in \mathbb{R}^2: \bm{A}\bm{v} \leq \bm{b} \}, \\
    \bm{A} = 
    \left[\begin{array}{c}
        (\bm{N}_{\text{R},1}^{\text{o}_i})^T \\
        (\bm{N}_{\text{R},2}^{\text{o}_i})^T
    \end{array}\right] \in \mathbb{R}^{2 \times 2}, \,
    \bm{b} = 
    \left[\begin{array}{c}
        c_{\text{R}, 1}^{\text{o}_i} \\
        c_{\text{R}, 2}^{\text{o}_i}
    \end{array}\right] \in \mathbb{R}^2.
    \label{eq:project_con1}
\end{gathered}
\end{equation*}
In this scenario, we need to project $\bm{v}_\text{R}$ onto the nearest hyperplane with respect to $\bm{A}$ and $\bm{b}$ for safety, as shown in Fig.~\ref{fig:vo_project}.
Then the GeoPro-VO can be described by
\begin{equation*}
    \mathcal{P}_{\text{VO}_i}(\bm{v}_\text{R}) = \bm{v}_\text{R} - \frac{\bm{a} (\bm{a}^T\bm{v}_\text{R} - b) }{\left\| \bm{a} \right\|_2},
\end{equation*}
where $\bm{a}^T \bm{x} = b$ represents the closest hyperplane for projecting $\bm{v}_\text{R}$ w.r.t $\bm{A}$ and $\bm{b}$.

In conclusion, with our proposed GeoPro-VO, we can project the unsafe velocity into the safe range to achieve dynamic obstacle avoidance between robot and obstacles.
In addition, physical limitations of states such as $v_x^{\text{min}} \leq v_x \leq v_x^{\text{max}}$ and $v_y^{\text{min}} \leq v_y \leq v_y^{\text{max}}$ can also be described by GeoPro, as
\begin{equation}
    \mathcal{P}_{\mathcal{D}_{\bm{x}}}(\bm{v}_\text{R}) = 
    \left\{
    \begin{aligned}
        & \bm{v}_\text{R}, \bm{v}^\text{min} \preceq \bm{v}_\text{R} \preceq \bm{v}^\text{max} \\
        & \bm{v}^\text{min}, \bm{v}_\text{R} \preceq \bm{v}^\text{min} \\
        & \bm{v}^\text{max}, \bm{v}^\text{max} \preceq \bm{v}_\text{R}
    \end{aligned}
    \right..
    \label{eq:physical_project}
\end{equation}
By combining the proposed $\mathcal{P}_{\text{VO}_i}$ and $\mathcal{P}_{\mathcal{D}_{\bm{x}}}$ with the augmented Lagrangian framework, we can reformulate \eqref{eq:vo_mpc} into the form like \eqref{eq:lagrangian2}.
Physical limitations of controls are considered in solving \eqref{eq:lagrangian2} through spectral projected gradient descent (SPG), more details please refer to Sec.~\ref{sec:alspg}.
We need to mention through leveraging the projection feature of VO to construct GeoPro, we can eliminate integer variables in \eqref{eq:integer_cons1} and obtain real time solutions for practical applications.

\subsection{Augmented Lagrangian Spectral Projected Gradient Descent (ALSPG)}
\label{sec:alspg}
Since SPG by itself is inadequate for tackling robotics problems with complex nonlinear constraints, ALSPG is proposed as a solution to this issue.
Using the augmented Lagrangian framework, \eqref{eq:nmpc} is reformulated as \eqref{eq:lagrangian2}.
It should be noted that \eqref{eq:states_cons} is excluded in \eqref{eq:lagrangian2} for brevity; however, the physical limitations of states should also be included in \eqref{eq:lagrangian2} in practice.
For convenience of updating the penalty parameters, we define a distance function as follows:
\begin{equation}
    V_{\mathcal{C}_i}(\mathcal{X}, \bm{\lambda}_{\mathcal{C}_i}, \rho_{\mathcal{C}_i}) := \bm{g}_i(\mathcal{X}) - \mathcal{P}_{\mathcal{C}_i}(\bm{g}_i(\mathcal{X}) + \frac{\bm{\lambda}_{\mathcal{C}_i}}{\rho_{\mathcal{C}_i}}).
    \label{eq:terminal_condition}
\end{equation}
Furthermore, let $V := [V_{\mathcal{C}_1}^T, V_{\mathcal{C}_2}^T, \dots, V_{\mathcal{C}_{N_p}}^T]^T \in \mathbb{R}^{\sum_{i=1}^{N_p} n_i}$.
The norm of $V$ can be utilized to design a termination criterion of ALSPG, when $\| V \| \leq \varepsilon_{\text{tol}}$, $\varepsilon_{\text{tol}} > 0$, iterations of ALSPG will terminate.
Since \eqref{eq:nmpc} includes an additional system constraint \eqref{eq:system_cons}, we reformulate \eqref{eq:lagrangian2} by eliminating the variable $\mathcal{X}$ using \eqref{eq:system_cons}.
First, through the standard linearization of the system dynamics \eqref{eq:system_cons}, we can derive that
\begin{equation*}
    A_k = \frac{\partial f(\bm{x}_k, \bm{u}_k)}{\partial \bm{x}_k}, \, B_k = \frac{\partial f(\bm{x}_k, \bm{u}_k)}{\partial \bm{u}_k},
\end{equation*}
where $A_k \in \mathbb{R}^{n_x \times n_x}$ and $B_k \in \mathbb{R}^{n_x \times n_u}$.
Moreover, the system dynamics integration is represented as $\mathcal{X} = \mathcal{A}\bm{x}_0 + \mathcal{B}\mathcal{U}$:
\begin{equation*}
    \resizebox{1.0\linewidth}{!}{$
    \mathcal{A} = 
    \left[\begin{array}{c}
    A_0 \\
    A_1A_0 \\
    \vdots \\
    \prod_{i=0}^{N-2}A_i \\
    \prod_{i=0}^{N-1}A_i 
    \end{array}\right], \, 
    \mathcal{B} = 
    \left[\begin{array}{ccccc}
    B_0 & 0 & \cdots & 0 & 0 \\
    A_1B_0 & B_1 & \cdots & 0 & 0 \\
    \vdots & \vdots & \ddots & \vdots & \vdots \\
    \prod_{i=1}^{N-2}A_iB_0 & \prod_{i=2}^{N-2}A_iB_1 & \dots & B_{N-2} & 0 \\
    \prod_{i=1}^{N-1}A_iB_0 & \prod_{i=2}^{N-1}A_iB_1 & \dots & A_{N-1}B_{N-2} & B_{N-1}
    \end{array}\right]. 
    $}
\end{equation*}
where $\mathcal{A} \in \mathbb{R}^{(N \cdot n_x) \times n_x}$, $\mathcal{B} \in \mathbb{R}^{(N \cdot n_x) \times (N \cdot n_u)}$ and $\bm{x}_0$ is the initial state of the robot at time step 0.
Let $\phi(\mathcal{U}):= \{\mathcal{X} \in \mathbb{R}^{N \cdot n_x}: \mathcal{A}\bm{x}_0 + \mathcal{B}\mathcal{U} \}$.
In addition, the cost function of \eqref{eq:nmpc} can be rewritten as $l(\mathcal{X}, \mathcal{U}) := l(\phi(\mathcal{U}), \mathcal{U}) = l(\mathcal{U})$.
With the above equations, \eqref{eq:lagrangian2} is simplified to
\begin{equation}
\begin{aligned}
    \mathcal{L}(&\mathcal{U}, \bm{\lambda}, \bm{\rho}) := l(\mathcal{U}) + \\
    & \sum_{i=1}^{N_p} \frac{\rho_{\mathcal{C}_i}}{2} \| \bm{g}_i(\mathcal{U}) + \frac{\bm{\lambda}_{\mathcal{C}_i}}{\rho_{\mathcal{C}_i}} - \mathcal{P}_{\mathcal{C}_i}(\bm{g}_i(\mathcal{U}) + \frac{\bm{\lambda}_{\mathcal{C}_i}}{\rho_{\mathcal{C}_i}}) \|^2.
    \label{eq:lagrangian3}
\end{aligned}
\end{equation}
And \eqref{eq:terminal_condition} is also simplified to:
\begin{equation*}
    V_{\mathcal{C}_i}(\mathcal{U}, \bm{\lambda}_{\mathcal{C}_i}, \rho_{\mathcal{C}_i}) := \bm{g}_i(\mathcal{U}) - \mathcal{P}_{\mathcal{C}_i}(\bm{g}_i(\mathcal{U}) + \frac{\bm{\lambda}_{\mathcal{C}_i}}{\rho_{\mathcal{C}_i}}).
\end{equation*}
The Jacobi matrices of the cost function $l(\mathcal{U})$ w.r.t $\mathcal{X}$ and $\mathcal{U}$ are represented by $\bm{J}_{\mathcal{X}} = \frac{\partial l(\mathcal{X}, \, \mathcal{U})}{\partial \mathcal{X}} \in \mathbb{R}^{N \cdot n_x}$ and $\bm{J}_{\mathcal{U}} = \frac{\partial l(\mathcal{X}, \, \mathcal{U})}{\partial \mathcal{U}} \in \mathbb{R}^{N \cdot n_u}$.
Thus the partial derivatives of $l(\mathcal{U})$ and $\bm{g}_i(\mathcal{U})$ w.r.t $\mathcal{U}$ are:
\begin{equation*}
\begin{aligned}
    & \triangledown l(\mathcal{U}) = \frac{\partial l(\mathcal{U})}{\partial \mathcal{U}} + \frac{\partial \phi(\mathcal{U})}{\partial \mathcal{U}} \frac{\partial l(\phi(\mathcal{U}))}{\partial \phi(\mathcal{U})} 
    = \bm{J}_\mathcal{U} + \mathcal{B}^T \bm{J}_\mathcal{X} \\
    & \frac{\partial \bm{g}_i(\mathcal{U})}{\partial \mathcal{U}} = \frac{\partial \phi(\mathcal{U})}{\partial \mathcal{U}} \frac{\partial \bm{g}_i(\phi(\mathcal{U}))}{\partial \phi(\mathcal{U})} = \mathcal{B}^T \triangledown \bm{g}_i
\end{aligned}
\end{equation*}
where $\triangledown \bm{g}_i = \frac{\partial \bm{g}_i(\mathcal{X})}{\partial \mathcal{X}} \in \mathbb{R}^{N \cdot n_x \times n_i}$.
Moreover, the derivative of \eqref{eq:lagrangian3} with respective to time is:
\begin{equation}
\begin{aligned}
    \triangledown \mathcal{L}(\mathcal{U}, \bm{\lambda}, \bm{\rho}) &= \triangledown l(\mathcal{U}) + \sum_{i=1}^{N_p} \rho_{\mathcal{C}_i} \frac{\partial \bm{g}_i(\mathcal{U})}{\partial \mathcal{U}} (V_{\mathcal{C}_i} + \frac{\bm{\lambda}_{\mathcal{C}_i}}{\rho_{\mathcal{C}_i}}) \\
    &= \mathcal{B}^T\bm{J}_\mathcal{X} + \bm{J}_\mathcal{U} + \sum_{i=1}^{N_p} \rho_{\mathcal{C}_i} \mathcal{B}^T \triangledown \bm{g}_i (V_{\mathcal{C}_i} + \frac{\bm{\lambda}_{\mathcal{C}_i}}{\rho_{\mathcal{C}_i}}) \\
    &= \mathcal{B}^T(\bm{J}_\mathcal{X} + \triangledown \mathcal{G} \Lambda \mathcal{V}) + \bm{J}_\mathcal{U},
\end{aligned}
\label{eq:lagrangian_derive}
\end{equation}
where
\begin{equation*}
\begin{aligned}
    \triangledown \mathcal{G} &= 
    \left[\begin{array}{cccc}
        \triangledown g_1 & \triangledown g_2 & \cdots & \triangledown g_{N_p}
    \end{array}\right] \in \mathbb{R}^{N \cdot n_x \times \sum_{i=1}^{N_p} n_i} \\
    \Lambda &= 
    \begin{bmatrix}  
        \rho_{\mathcal{C}_1}I & & & \\  
        & \rho_{\mathcal{C}_2} I & & \\  
        & & \ddots & \\  
        & & & \rho_{\mathcal{C}_{N_p}} I  
    \end{bmatrix} \in \mathbb{R}^{\sum_{i=1}^{N_p} n_i \times \sum_{i=1}^{N_p} n_i} \\
    & \rho_{\mathcal{C}_i}I = 
    \begin{bmatrix}  
        \rho_{\mathcal{C}_i} & 0 & \cdots & 0 \\  
        0 & \rho_{\mathcal{C}_i} & \cdots & 0 \\  
        \vdots & \vdots & \ddots & \vdots \\  
        0 & 0 & \cdots & \rho_{\mathcal{C}_i}  
    \end{bmatrix} \in \mathbb{R}^{n_i \times n_i}, \\
    & \mathcal{V} = 
    \begin{bmatrix}
        V_{\mathcal{C}_1} + \frac{\bm{\lambda}_{\mathcal{C}_1}}{\rho_{\mathcal{C}_1}} \\
        V_{\mathcal{C}_2} + \frac{\bm{\lambda}_{\mathcal{C}_2}}{\rho_{\mathcal{C}_2}} \\
        \vdots \\
        V_{\mathcal{C}_{N_p}} + \frac{\bm{\lambda}_{\mathcal{C}_{N_p}}}{\rho_{\mathcal{C}_{N_p}}}
    \end{bmatrix} \in \mathbb{R}^{\sum_{i=1}^{N_p} n_i}
\end{aligned}
\end{equation*}
To efficiently compute \eqref{eq:lagrangian_derive}, we use a recursive iteration approach.
Let $\bm{\omega} = [\bm{\omega}_0^T, \bm{\omega}_1T, \dots, \bm{\omega}_{N - 1}^T]^T \in \mathbb{R}^{N \cdot n_x}$ be the vector to multiply, and $\bm{z} = [\bm{z}_0^T, \bm{z}_1^T, \dots, \bm{z}_{N - 1}^T]^T \in \mathbb{R}^{N \cdot n_u} $ be the resulting vector.
Then we have $\bm{z} = \mathcal{B}^T \bm{\omega}$:
\begin{equation*}
\resizebox{1.0\linewidth}{!}{$
\begin{bmatrix}
    \bm{z}_0 \\
    \bm{z}_1 \\
    \vdots \\
    \bm{z}_{N - 2} \\
    \bm{z}_{N - 1}
\end{bmatrix} = 
\begin{bmatrix}
    B_0^T & B_0^T A_1^T & \cdots & B_0^T \prod_{i=1}^{N-2} A_i^T & B_0^T \prod_{i=1}^{N-1} A_i^T \\
    0 & B_1^T & \cdots & B_1^T \prod_{i=2}^{N-2} A_i^T & B_1^T \prod_{i=2}^{N-1} A_i^T \\
    \vdots & \vdots & \ddots & \vdots & \vdots \\
    0 & 0 & \cdots & B_{N-2}^T & B_{N-2}^T A_{N-1}^T \\
    0 & 0 & \cdots & 0 & B_{N-1}^T 
\end{bmatrix}
\begin{bmatrix}
    \bm{\omega}_0 \\
    \bm{\omega}_1 \\
    \vdots \\
    \bm{\omega}_{N - 2} \\
    \bm{\omega}_{N - 1}
\end{bmatrix}.
$}
\end{equation*}
In addition, we have the following recursive equations in a reverse order:
\begin{equation*}
\begin{aligned}
    &\bm{z}_k = B_k^T \widetilde{\bm{z}}_k, \\
    &\widetilde{\bm{z}}_k = \bm{\omega}_k + A_{k+1}^T\widetilde{\bm{z}}_{k+1}, \\
    &\widetilde{\bm{z}}_{N-1} = \bm{w}_{N-1}.
\end{aligned}
\end{equation*}

To conclusion, with the augmented Lagrangian framework we have reformulated \eqref{eq:nmpc} as a constrained and non-convex problem:
\begin{equation}
    \mathop{\arg\min}\limits_{\mathcal{U} \in \mathcal{D}_u} \mathcal{L}(\mathcal{U}, \bm{\lambda}, \bm{\rho}).
    \label{eq:lagrangian4}
\end{equation}
Moreover, we can solve this sub optimization-problem \eqref{eq:lagrangian4} using SPG with the geometric projector $\mathcal{P}_{\mathcal{D}_u}$ that is similar to \eqref{eq:physical_project}.
For more details about SPG, please refer to~\cite{birgin2014spectral}.
The detailed steps of ALSPG are outlined in Alg.~\ref{algorithm1}, where $\mathcal{P}_{\mathcal{C}_i}$ encompasses both $\mathcal{P}_{\text{VO}_i}$ and $\mathcal{P}_{\mathcal{D}_x}$.
After obtaining the optimal control $\mathcal{U}_\text{opt}$, we apply the first element to update the robot's state.
We then rerun Alg.\ref{algorithm1} until the robot reaches its destination.

In summarize, through combing GeoPro-VO and ALSPG, the MINLP~\eqref{eq:vo_mpc} is converted to a constrained problem like~\eqref{eq:lagrangian4}, which can greatly improve computational efficiency and obtain real time solution in practice.
\begin{algorithm}
    \caption{ALSPG}
    \label{algorithm1}
    \KwIn{$\bm{x}_0$, $\mathcal{U}_\text{init}$, $\mathcal{P}$: $\mathcal{P}_\text{VO}$, $\mathcal{P}_{\mathcal{D}_x}$, $\mathcal{P}_{\mathcal{D}_u}$, $\bm{\lambda}_{\mathcal{C}_i}=\bm{0}$, $\rho_{\mathcal{C}_i}=0.1$, $\beta=20$, $k=0$, $N_{\text{iter}}=0$, $N_\text{max}$, $\varepsilon_{\text{tol}}=1^{-2}$}
    \While{$N_\text{iter} \leq N_\text{max}$}{
        $\mathcal{U}_{k+1} = \mathop{\arg\min}\limits_{\mathcal{U}_k \in \mathcal{D}_u} \mathcal{L}(\mathcal{U}_k, \bm{\lambda}^k, \bm{\rho}^k)$ with SPG \;
        \For{$\mathcal{P}_{\mathcal{C}_i}$}{
            $\bm{\lambda}_{\mathcal{C}_i}^{k+1} = \rho_{\mathcal{C}_i}^k (V_{\mathcal{C}_i}(\mathcal{U}_{k+1}, \bm{\lambda}_{\mathcal{C}_i}^k, \rho_{\mathcal{C}_i}^k) + \frac{\bm{\lambda}_{\mathcal{C}_i}^k}{\rho_{\mathcal{C}_i}^k})$ \;
            \eIf{{\small$V_{\mathcal{C}_i}(\mathcal{U}_{k+1}, \bm{\lambda}_{\mathcal{C}_i}^{k+1}, \rho_{\mathcal{C}_i}^k) \leq V_{\mathcal{C}_i}(\mathcal{U}_k, \bm{\lambda}_{\mathcal{C}_i}^k, \rho_{\mathcal{C}_i}^k)$}}
            {$\rho_{\mathcal{C}_i}^{k+1} = \rho_{\mathcal{C}_i}^k$\;}{$\rho_{\mathcal{C}_i}^{k+1} = \beta\rho_{\mathcal{C}_i}^k$\;}
        }
        \If{$\| V \| \leq \varepsilon_{tol}$}{break\;}
        $N_{\text{iter}} \leftarrow N_{\text{iter}} + 1$\;
    }
    \KwOut{Optimal control $\mathcal{U}_\text{opt}$.}
\end{algorithm}

\section{Numerical Simulations}
\label{sec:experiments}
\subsection{Implementation Details}
\begin{table}
    \caption{Setup of Simulation Parameters}
    \label{tab:simulation_params}
    \centering
    \begin{tabular}{l|l|l}
    \hline
    Notation & Meaning & Value     \\ \hline
    $v_\text{x}^{\text{max}}$ & Robot's maximum horizontal velocity & $0.4 \, \si[per-mode=symbol]{\metre\per\second}$ \\
    $v_\text{y}^{\text{max}}$ & Robot's maximum vertical velocity & $0.4 \, \si[per-mode=symbol]{\metre\per\second}$ \\
    $a_\text{x}^{\text{max}}$ & Robot's maximum horizontal acceleration & $1.0 \, \si[per-mode=symbol]{\metre\per\second^2}$ \\
    $a_\text{y}^{\text{max}}$ & Robot's maximum vertical acceleration & $1.0 \, \si[per-mode=symbol]{\metre\per\second^2}$ \\
    $r$ & Radius of robot & $0.1 \, \si[per-mode=symbol]{\metre}$ \\
    $d_\text{s}$ & Safe margin for collision avoidance & $0.03 \, \si[per-mode=symbol]{\metre}$ \\
    $\epsilon_\text{tol}$ & Termination criterion of ALSPG & $1^{-2}$ \\
    $N_\text{max}$ & Maximum number of iterations & 20 \\
    $N$ & Prediction horizon of NMPC & 6 \\ 
    $\beta$ & Penalty factor of ALSPG & 20 \\
    $\Delta t$ & Time step of simulation & $0.05 \, \si[per-mode=symbol]{\second}$ \\
    \hline     \end{tabular}%
\end{table}
The effectiveness and performance of GeoPro-VO are evaluated through numerical simulations.
All simulations are conducted on an Ubuntu Laptop with Intel Core i9-13900HX processor using Python for all computations.
The simulation time step $\Delta t$ is set as $0.05 \, \si[per-mode=symbol]{\second}$.
Furthermore, the maximum values for horizontal velocity $v_\text{x}^\text{max} = -v_\text{x}^\text{min}$ and acceleration $a_\text{x}^\text{max} = -a_\text{x}^\text{min}$, as well as for vertical velocity $v_\text{y}^\text{max} = -v_\text{y}^\text{min}$ and acceleration $a_\text{y}^\text{max} = -a_\text{y}^\text{min}$, represent the physical limitations of the double integral model \eqref{eq:robot_dynamics}.
To meet safety requirements, we incorporate a safe margin $d_\text{s}$ into the radius of robot.
The parameters for both the robot dynamics and ALSPG are shown in Tab.~\ref{tab:simulation_params}.
In the following, we will present simulation results and compare our method with the state of the art.

\subsection{Simulation Results with Static and Dynamic Obstacles}
\begin{figure}
    \centering
    \includegraphics[width=0.98\linewidth]{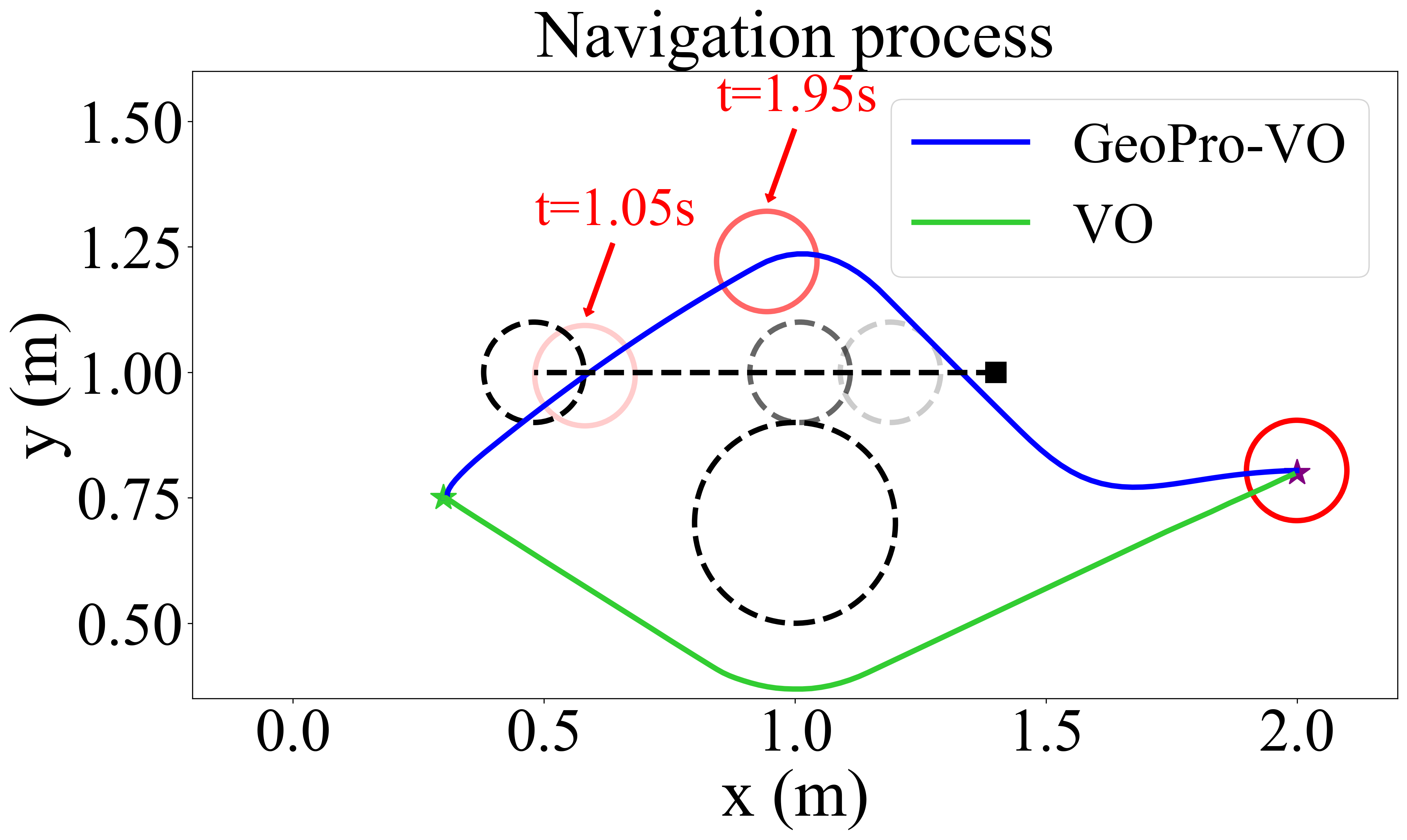}
    \caption{Robot navigation process}
    \caption{Simulation results of navigating the robot to its destination while considering both static and dynamic obstacles. The trajectories of our methods (GeoPro-VO) and VO are depicted by blue and green lines, respectively. The starting and ending points of the robot are marked with green and purple stars, respectively. The robot is represented by red circles, while obstacles are shown as black dashed circles. The positions of the robot and dynamic obstacle at various moments are depicted using varying color depth.}  
    \label{fig:navigation}  
\end{figure}
\begin{figure}
    \centering
    \begin{subfigure}{0.49\linewidth}
        \centering
        \includegraphics[width=0.98\linewidth]{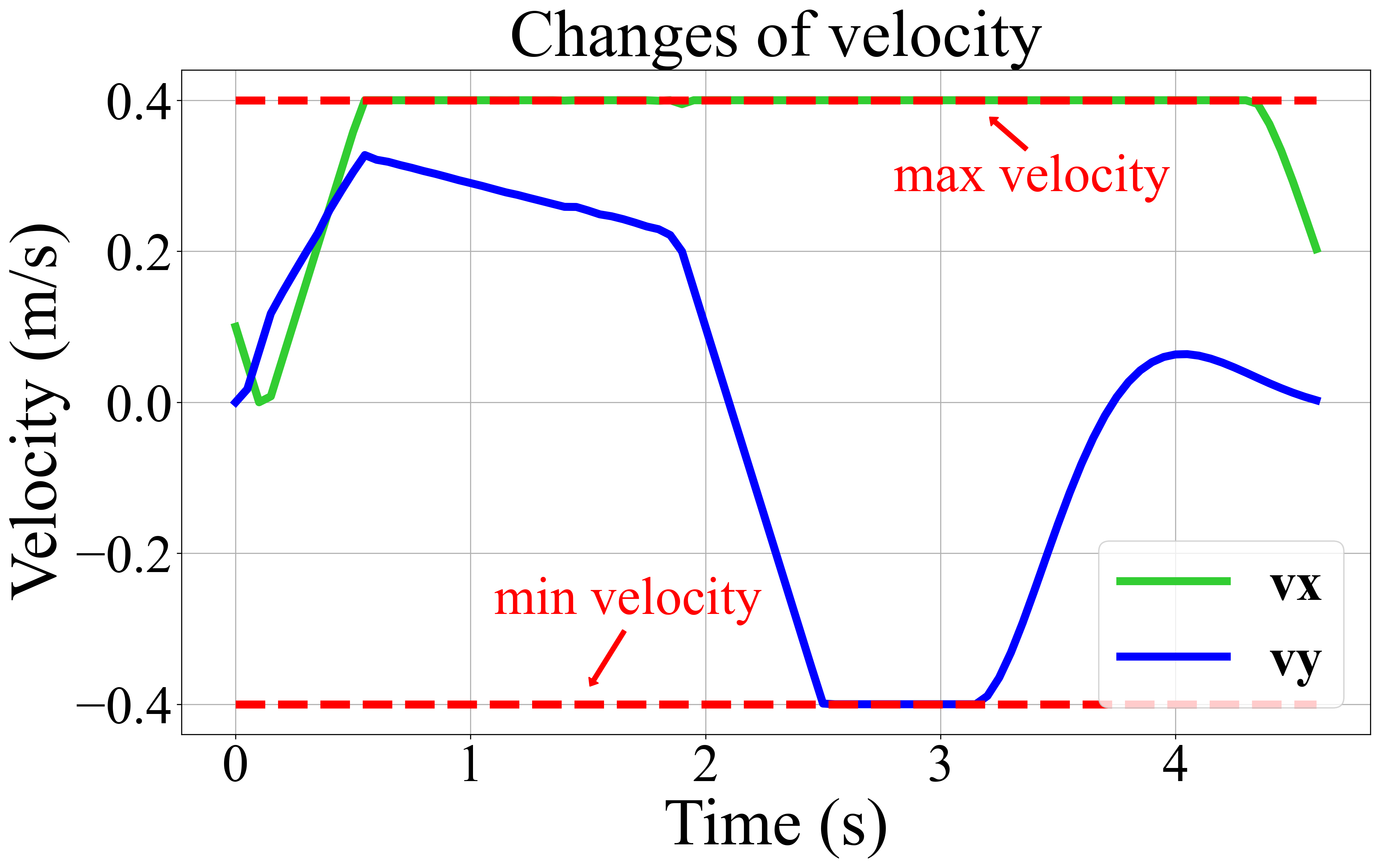}
        \caption{Velocity changes of GeoPro-VO}
        \label{subfig:states_case1}
    \end{subfigure}
    \centering
    \begin{subfigure}{0.49\linewidth}
        \centering
        \includegraphics[width=0.98\linewidth]{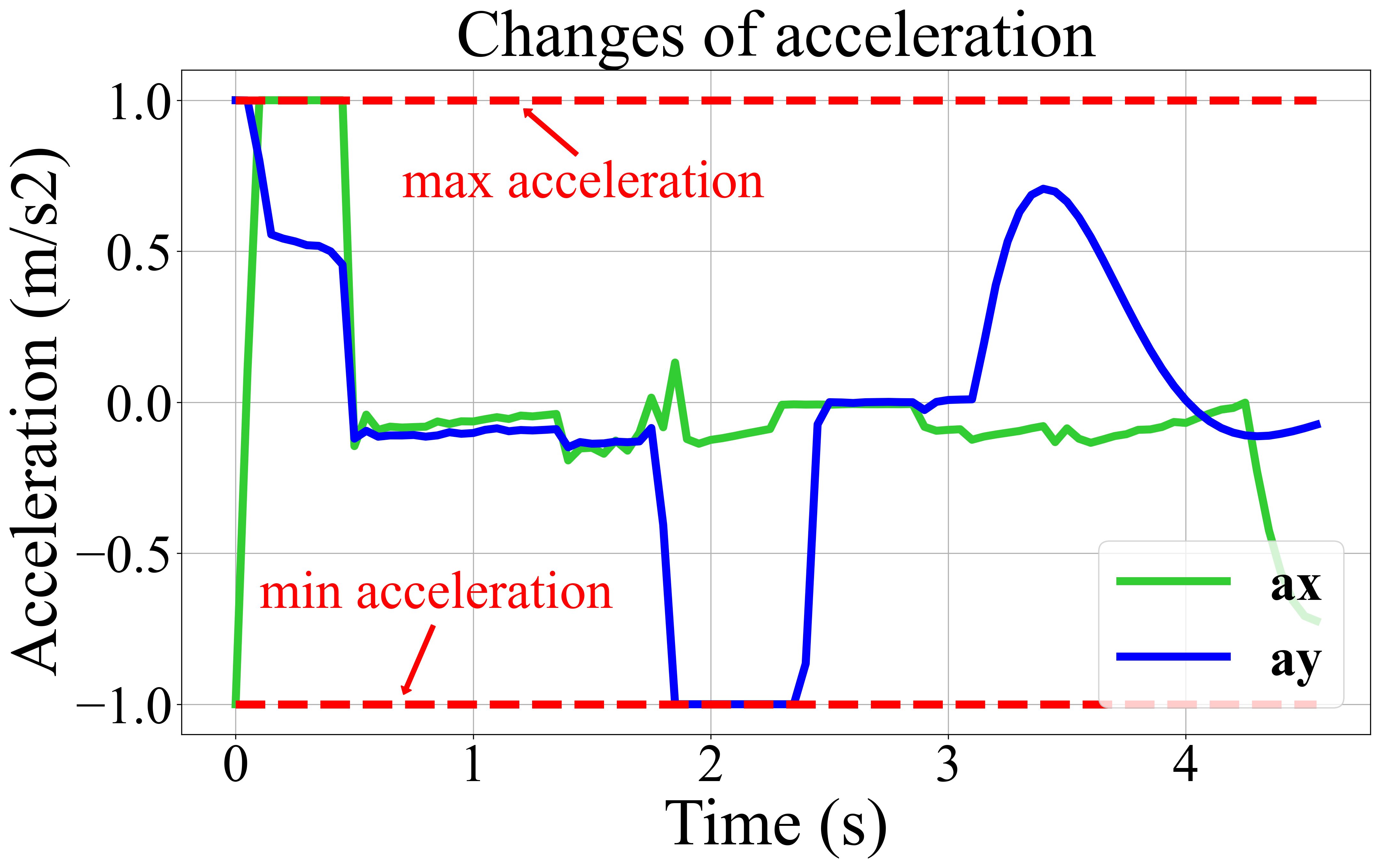}
        \caption{Acceleration changes of GeoPro-VO}
        \label{subfig:controls_case1}
    \end{subfigure}

    \centering
    \begin{subfigure}{0.49\linewidth}
        \centering
        \includegraphics[width=0.98\linewidth]{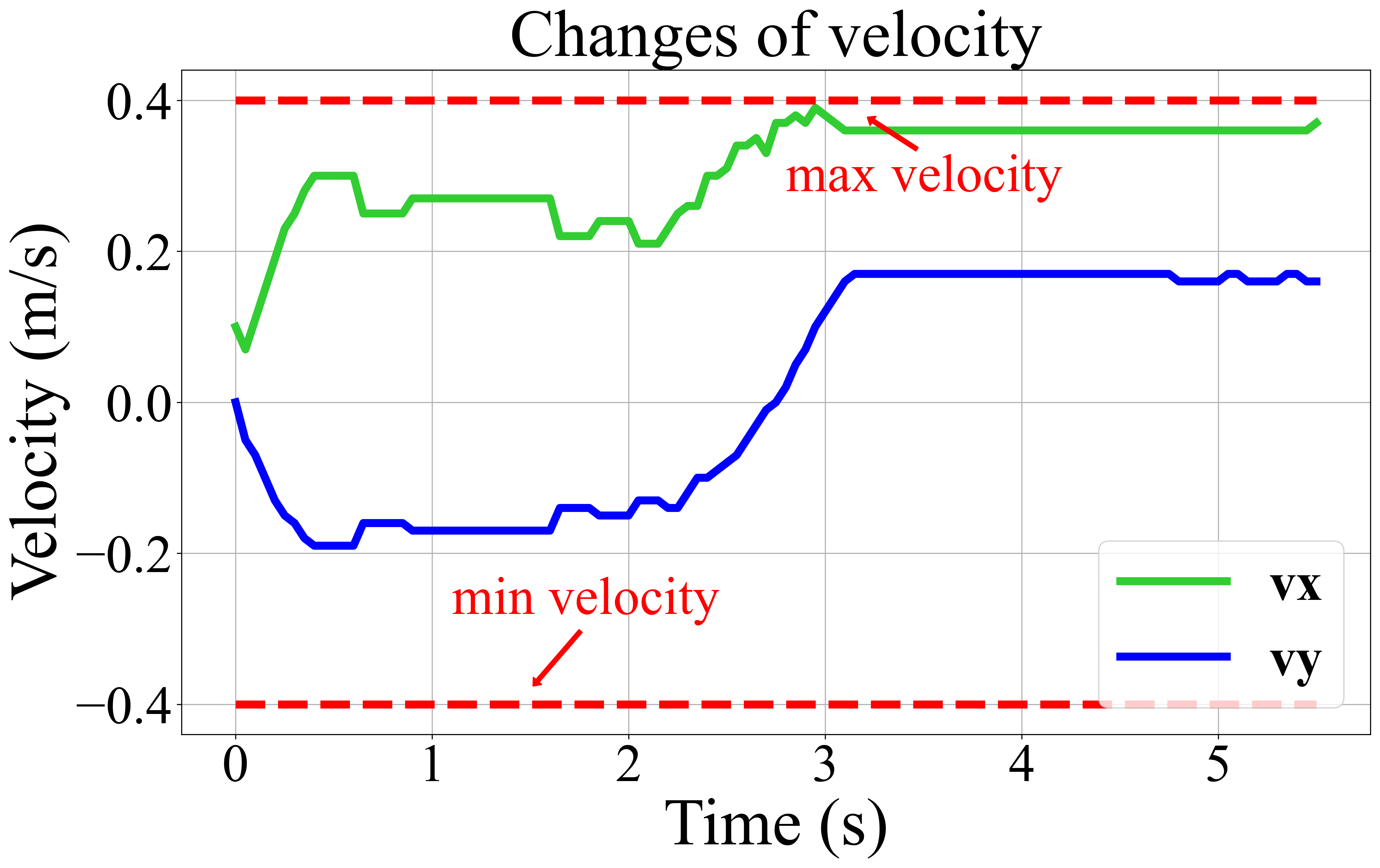}
        \caption{Velocity changes of VO}
        \label{subfig:states_case2}
    \end{subfigure}
    \centering
    \begin{subfigure}{0.49\linewidth}
        \centering
        \includegraphics[width=0.98\linewidth]{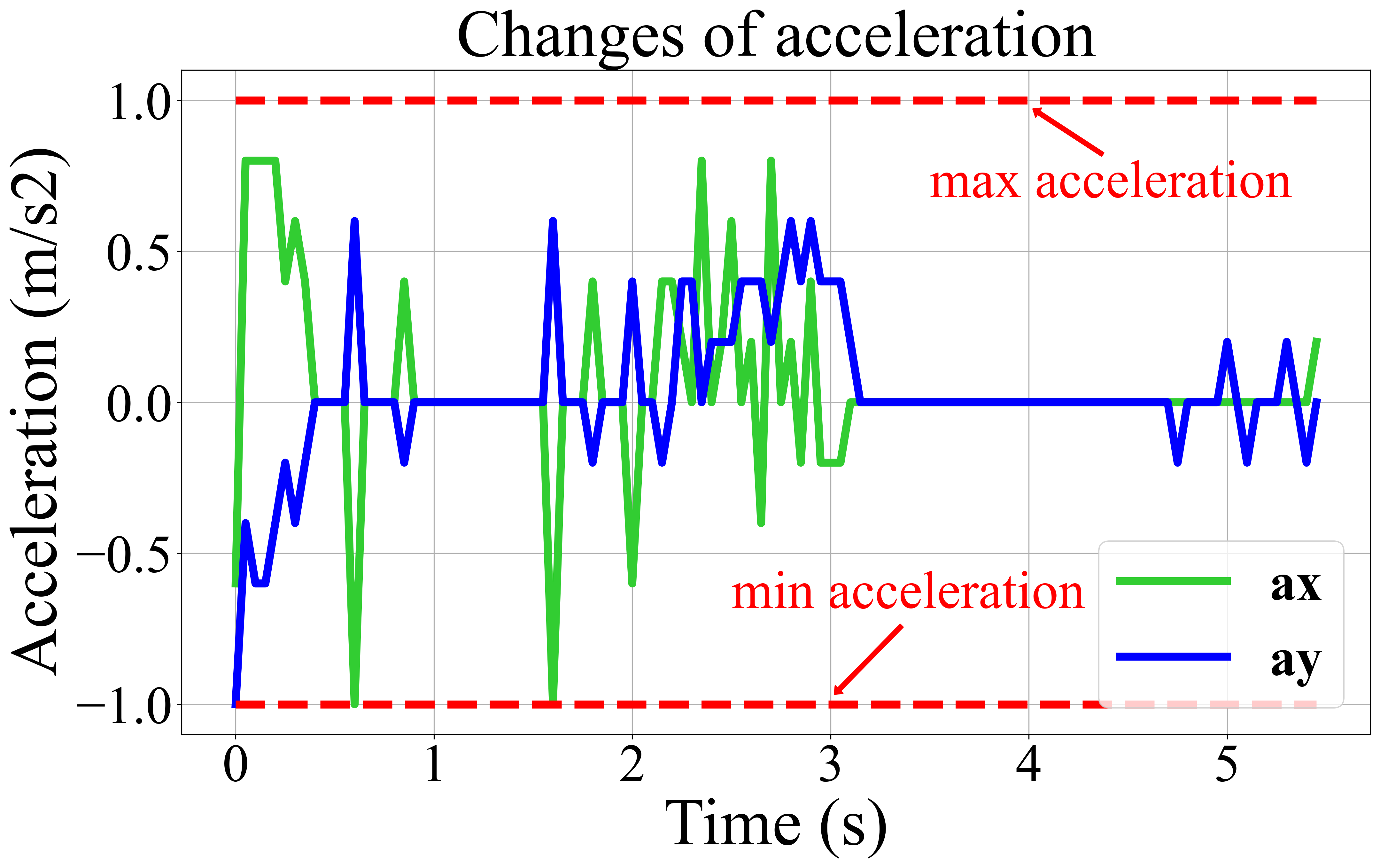}
        \caption{Acceleration changes of VO}
        \label{subfig:controls_case2}
    \end{subfigure}
    \caption{Velocity and acceleration changes of our method GeoPro-VO and VO.} 
    \label{fig:physical_cons}
\end{figure}
In this section, we mainly demonstrate effectiveness of our method.
The prediction horizon of NMPC is set as $6$.
The robot's initial and final positions are at $(0.3 \, \si[per-mode=symbol]{\metre}, 0.75 \, \si[per-mode=symbol]{\metre})$ and $(2.0 \, \si[per-mode=symbol]{\metre}, 0.8 \, \si[per-mode=symbol]{\metre})$, respectively.
Moreover, there are two obstacles in the environment: one static obstacle and another dynamic obstacle moved at a velocity $(-0.2 \, \si[per-mode=symbol]{\metre\per\second}, 0.0 \, \si[per-mode=symbol]{\metre\per\second})$.
The navigation process of the robot is illustrated in Fig.~\ref{fig:navigation}, our method successfully navigate the robot towards its destination while avoiding collisions with all obstacles, where the trajectory of our method is depicted in blue, while the trajectory of the nominal VO~\cite{fiorini1998motion} is shown in green, and the dynamic obstacle's path is represented by a black dashed line.
The nominal VO method does not utilize the constrained NMPC framework, hence it tends to avoid obstacles from below for safety requirements.
In contrast, our approach opts to evade obstacles on the same side as it also accounts for safety constraints within the prediction horizon.
In addition, when reducing $N$ to $2$, our method also chooses to avoid obstacles from below.
The positions of both the robot and dynamic obstacle at different time moments are indicated by varying colors' depths.

The velocity and acceleration change curves for both our method and the nominal VO are shown in Fig.~\ref{fig:physical_cons}, illustrating the changes during the navigation process.
It is evident that both approaches adhere to the physical limitations of states and controls.
However, since the nominal VO is designed for collision avoidance in the velocity space, it is unsuitable for a robot controlled by acceleration.
This can lead to unstable behavior of acceleration, as illustrated in Fig.~\ref{subfig:controls_case2}.
On the contrary, our method manipulates in the acceleration space, resulting in smoother curves, as shown in Fig.~\ref{subfig:controls_case1}.
    
\subsection{Efficient Computation of ALSPG}
In this section, we mainly demonstrate the advantage for computation speed of ALSPG through comparing it with the commercial second-order solvers.
All optimization problems are formulated as constrained NMPC and solved using various solvers.
For constrained NMPC using VO as obstacle avoidance constraints, the nominal optimization problem~\eqref{eq:vo_mpc} is solved by BONMIN~\cite{sutradhar2016minlp}, which is a solver designed for MINLP.
Similarly, for constrained NMPC using Euclidean distance as obstacle avoidance constraints, the nominal optimization problem~\eqref{eq:nmpc} is solved by IPOPT~\cite{Biegler2009large}.
Furthermore, the constrained NMPC problems with two different types of constraints mentioned above can also be solved using ALSPG with the help of GeoPro.
The maximum computation time of ALSPG will occur in the first few runs of it because it requires more iterations to converge to assure that all constraints are satisfied.
To reduce the computation time, we can use the final result from the previous optimization round as the initial value for the next round.
This trick is also applicable to the commercial solvers.

The comparison results are shown in Tab.~\ref{tab:time_comparison}, where five scenarios are considered:
S2: a robot with two static obstacles; S4: a robot with four static obstacles; D1: a robot with one dynamic obstacle;
D2: a robot with two dynamic obstacles; D3: a robot with three dynamic obstacles.
We can observe that although IPOPT is implemented in C++, ALSPG (implemented in Python) has similar calculation speeds as it.
The increased efficiency is attributed to GeoPro's capability to handle non-smooth safe constraints like \eqref{eq:Euclidean}, as well as ALSPG only requires the first-order information of the objective function.
When comparing ALSPG to BONMIN, we can observe ALSPG also demonstrates greater computational efficiency.
The large computation time of VO-NMPC~\eqref{eq:vo_mpc} solved by BONMIN arises from it adds integer variables to the optimization problem~\eqref{eq:vo_mpc} to cope with the problem that at least one of two constraints~\eqref{eq:vo_cons2} needs to be satisfied, transforming the nominal problem \eqref{eq:nmpc} into a MINLP~\eqref{eq:vo_mpc}.

\subsection{Reliable Security of GeoPro-VO}
\begin{figure}
    \centering
    \begin{subfigure}{0.49\linewidth}
        \centering
        \includegraphics[width=0.98\linewidth]{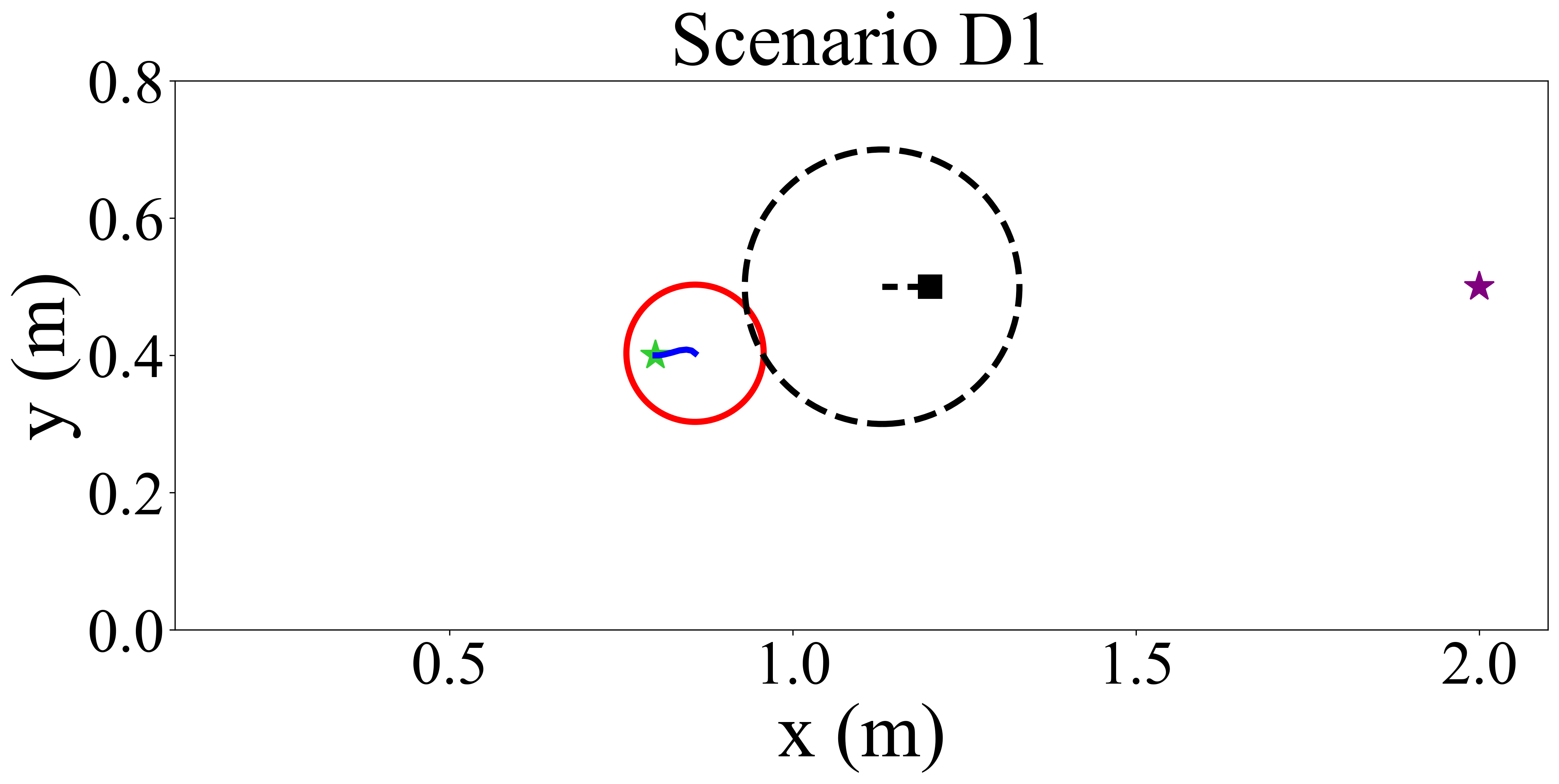}
        \caption{Scenario D1}
        \label{subfig:failD1}
    \end{subfigure}
    \centering
    \begin{subfigure}{0.49\linewidth}
        \centering
        \includegraphics[width=0.98\linewidth]{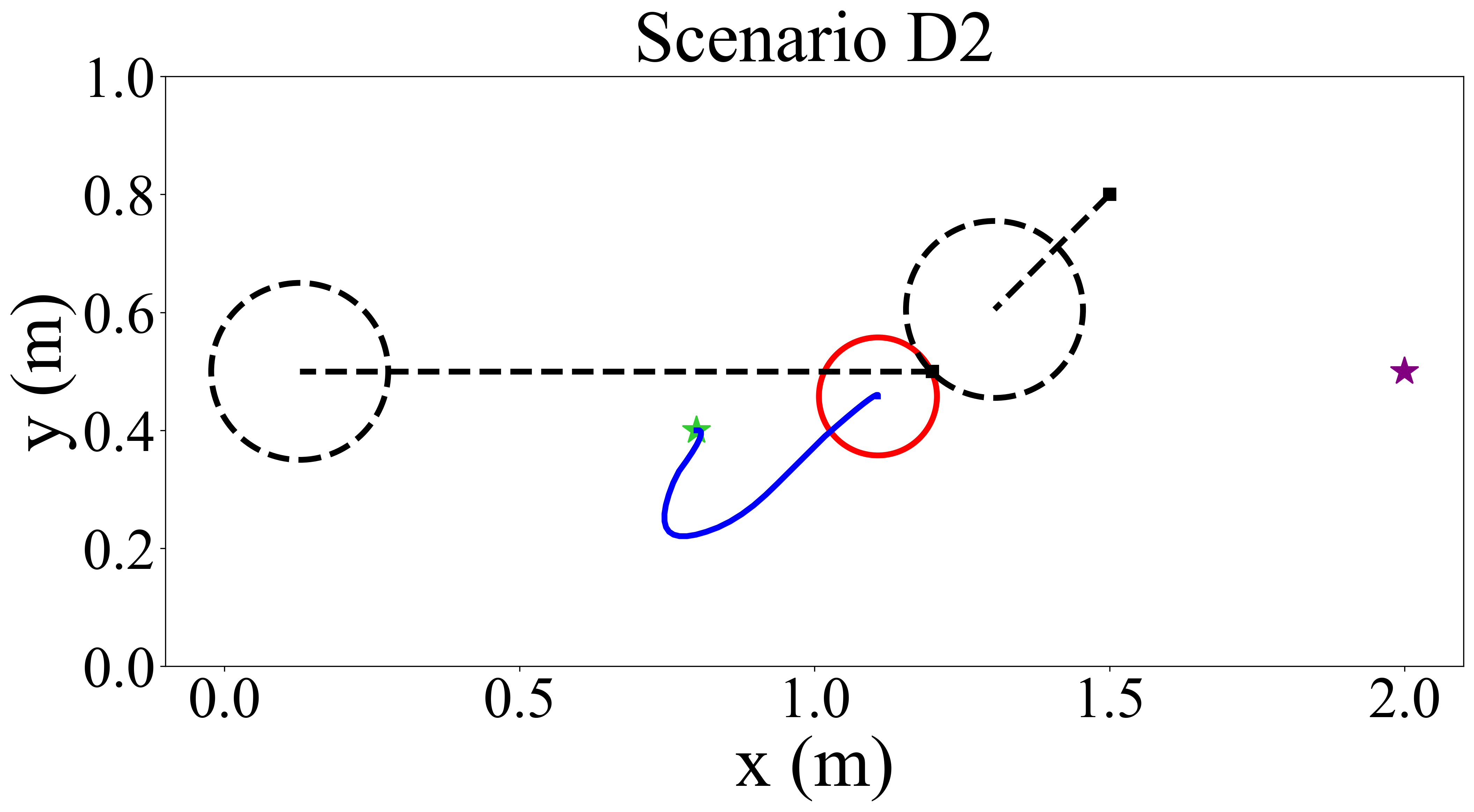}
        \caption{Scenario D2}
        \label{subfig:failD2}
    \end{subfigure}
    \caption{Failure scenarios when using ALSPG with GeoPro-ED for dynamic obstacle avoidance.
    (a): The prediction horizon of NMPC is set as 2.
    (b): The prediction horizon of NMPC is set as 6.} 
    \label{fig:failure_cases}
\end{figure}

\begin{table}[t]
\caption{Comparison of Different Types of Constraints in Optimization Problems}
\label{tab:time_comparison}
\centering
\resizebox{\columnwidth}{!}{%
\begin{threeparttable}
\begin{tabular}{ccccccc}
\hline
\multirow{2}{*}{\begin{tabular}[c]{@{}c@{}}Constraint\\ Types\tnote{a} \end{tabular}} &
  \multirow{2}{*}{Solver} &
  \multirow{2}{*}{Scenario} &
  \multicolumn{4}{c}{Time (ms)} \\ \cline{4-7} 
 &
   &
   &
  Max &
  Min &
  Median &
  Avg \\ \hline
\multirow{15}{*}{\begin{tabular}[c]{@{}c@{}}Velocity Obstacle\\ based Constraints\end{tabular}} &
  \multirow{5}{*}{\begin{tabular}[c]{@{}c@{}}ALSPG\tnote{b} \\ N=2\end{tabular}} &
  S2\tnote{e} &
  80.69 &
  2.86 &
  3.12 &
  14.29 \\
 &
   &
  S4 &
  200.57 &
  5.08 &
  49.42 &
  40.72 \\
 &
   &
  D1\tnote{f} &
  \textcolor{blue}{\textbf{84.08}} &
  \textcolor{blue}{\textbf{1.17}} &
  \textcolor{blue}{\textbf{1.76}} &
  \textcolor{blue}{\textbf{8.39}} \\
 &
   &
  D2 &
  \textcolor{blue}{\textbf{113.49}} &
  \textcolor{blue}{\textbf{2.83}} &
  \textcolor{blue}{\textbf{3.05}} &
  \textcolor{blue}{\textbf{10.97}} \\
 &
   &
  D3 &
  \textcolor{blue}{\textbf{121.56}} &
  \textcolor{blue}{\textbf{3.99}} &
  \textcolor{blue}{\textbf{8.78}} &
  \textcolor{blue}{\textbf{16.77}} \\ \cline{2-7} 
 &
  \multirow{5}{*}{\begin{tabular}[c]{@{}c@{}}ALSPG\\ N=6\end{tabular}} &
  S2 &
  126.01 &
  4.28 &
  89.41 &
  70.51 \\
 &
   &
  S4 &
  765.52 &
  10.22 &
  207.38 &
  236.48 \\
 &
   &
  D1 &
  130.29 &
  3.30 &
  52.81 &
  57.46 \\
 &
   &
  D2 &
  \textcolor{blue}{\textbf{435.53}} &
  \textcolor{blue}{\textbf{4.43}} &
  \textcolor{blue}{\textbf{86.49}} &
  \textcolor{blue}{\textbf{117.23}} \\
 &
   &
  D3 &
  \textcolor{blue}{\textbf{608.40}} &
  \textcolor{blue}{\textbf{6.31}} &
  \textcolor{blue}{\textbf{140.91}} &
  \textcolor{blue}{\textbf{157.28}} \\ \cline{2-7} 
 &
  \multirow{5}{*}{\begin{tabular}[c]{@{}c@{}}BONMIN\tnote{c} \\ N=6\end{tabular}} &
  S2 &
  2189.86 &
  63.64 &
  88.28 &
  287.81 \\
 &
   &
  S4 &
  157842 &
  111.18 &
  141.24 &
  9273.05 \\
 &
   &
  D1 &
  856.92 &
  43.33 &
  63.69 &
  114.27 \\
 &
   &
  D2 &
  3137.72 &
  63.53 &
  98.91 &
  377.28 \\
 &
   &
  D3 &
  4440.58 &
  85.46 &
  160.23 &
  947.50 \\ \hline
\multirow{15}{*}{\begin{tabular}[c]{@{}c@{}}Euclidean Distance\\ based Constraints\end{tabular}} &
  \multirow{5}{*}{\begin{tabular}[c]{@{}c@{}}ALSPG\\ N=2\end{tabular}} &
  S2 &
  15.15 &
  0.97 &
  1.08 &
  2.92 \\
 &
   &
  S4 &
  20.54 &
  1.36 &
  1.48 &
  3.96 \\
 &
   &
  D1 &
  N/A &
  N/A &
  N/A &
  N/A \\
 &
   &
  D2 &
  N/A &
  N/A &
  N/A &
  N/A \\
 &
   &
  D3 &
  N/A &
  N/A &
  N/A &
  N/A \\ \cline{2-7} 
 &
  \multirow{5}{*}{\begin{tabular}[c]{@{}c@{}}ALSPG\\ N=6\end{tabular}} &
  S2 &
  37.15 &
  1.19 &
  25.75 &
  20.15 \\
 &
   &
  S4 &
  51.81 &
  1.77 &
  37.26 &
  29.27 \\
 &
   &
  D1 &
  36.84 &
  1.15 &
  17.55 &
  16.46 \\
 &
   &
  D2 &
  N/A &
  N/A &
  N/A &
  N/A \\
 &
   &
  D3 &
  N/A &
  N/A &
  N/A &
  N/A \\ \cline{2-7} 
 &
  \multirow{5}{*}{\begin{tabular}[c]{@{}c@{}}IPOPT\tnote{d} \\ N=6\end{tabular}} &
  S2 &
  301.15 &
  12.77 &
  16.28 &
  21.23 \\
 &
   &
  S4 &
  301.44 &
  19.24 &
  22.81 &
  28.33 \\
 &
   &
  D1 &
  291.73 &
  10.98 &
  14.73 &
  24.15 \\
 &
   &
  D2 &
  N/A &
  N/A &
  N/A &
  N/A \\
 &
   &
  D3 &
  N/A &
  N/A &
  N/A &
  N/A \\ \hline
\end{tabular}%
\end{threeparttable}
}
\begin{tablenotes}
    \item [] \textsuperscript{a} All optimization setup are formulated as constrained NMPC problems.
    \item [] \textsuperscript{b} ALSPG is used in conjunction with the geometric projector.
    \item [] \textsuperscript{c} BONMIN is a solver for the mixed integer nonlinear programming problem.
    \item [] \textsuperscript{d} IPOPT is a solver for the nonlinear optimization problem.
    \item [] \textsuperscript{e} S2 means one robot and two static obstacles.
    \item [] \textsuperscript{f} D1 means one robot and one dynamic obstacle.
\end{tablenotes}
\end{table}
When comparing these two geometric projectors GeoPro-VO and GeoPro-ED, our method GeoPro-VO requires more computation time under the same prediction horizon, as shown in cases S2 and S4 in Tab.~\ref{tab:time_comparison}.
This is due to GeoPro-VO involves additional complex calculations for constructing the GeoPro, such as constructing hyperplanes and projecting the unsafe velocity onto the nearest hyperplane.
In contrast, GeoPro-ED involves fewer calculations with lower computational complexity, as shown in \eqref{eq:Euclid_pro}.
When analyzing scenarios with dynamic obstacles, we can observe that ALSPG with GeoPro-ED in the case of prediction horizon of $2$ fails to achieve collision avoidance with the dynamic obstacle in scenario D1, as shown in Fig.~\ref{subfig:failD1}.
Moreover, for the other two scenarios D2 and D3, ALSPG with GeoPro-ED also fails to avoid collisions with dynamic obstacles even with prediction horizon of $6$, the details of scenario D2 is shown in Fig.~\ref{subfig:failD2}.
In contrast, ALSPG with our method GeoPro-VO successfully avoids collisions with all dynamic obstacles even the MPC prediction horizon is set at $2$.
This difference arises from VO's ability to anticipate and avoid collisions in the velocity space, whereas GeoPro-ED focuses on collision avoidance in the position space.
In addition, since GeoPro-VO has the ability of assuring safety even with a short prediction of MPC, we can reduce the prediction horizon of MPC when solving through ALSPG with GeoPro-VO to improve solution efficiency.

In conclusion, solving VO-NMPC using ALSPG with GeoPro-VO is more computationally efficient than solved by BONMIN.
In comparison to GeoPro-ED, GeoPro-VO requires more time for computation due to the complexity involved in constructing the GeoPro.
However, GeoPro-VO performs well in ensuring safety with dynamic obstacles when solving the constrained NMPC problem with a short prediction horizon through ALSPG, while GeoPro-ED falls short in this aspect.

\section{Conclusions}
\label{sec:con}
In this paper, we propose a geometric projector for dynamic obstacle avoidance based on velocity obstacle (GeoPro-VO).
Furthermore, we integrate GeoPro-VO with the augmented Lagrangian spectral projected gradient descent (ALSPG) algorithm to reformulate the nominal constrained nonlinear model predictive control (NMPC) problem as a sub-optimization problem and solve it efficiently.
The effectiveness and performance of our method is validated through numerical simulations, and results demonstrate that our method successfully guides the robot to its destination without any collisions with both static and dynamic obstacles.
Compared to the start of the art, our method is more computationally efficient and can ensures safety with dynamic obstacles even with a short prediction horizon of NMPC.
Future research will focus on improving the computational efficiency of our method and extending it to multi-robot systems.

{
\bibliographystyle{IEEEtran}
\bibliography{reference}
}

\end{document}